\documentclass[11pt]{article}

\usepackage[final]{acl}

\usepackage{amsmath}

\usepackage{booktabs}

\usepackage{times}
\usepackage{latexsym}

\usepackage[T1]{fontenc}
\usepackage[utf8]{inputenc}

\usepackage{microtype}

\usepackage{inconsolata}

\usepackage{graphicx}

\title{Toward Agentic RAG for Ukrainian\thanks{This paper is a research report based on our participation in the UNLP 2026 Shared Task.}}

\author{
Marta Sumyk \and Oleksandr Kosovan \\
Ukrainian Catholic University, Lviv, Ukraine \\
\texttt{sumyk.pn@ucu.edu.ua, o.kosovan@ucu.edu.ua}
}

\begin{document}
\maketitle
\begin{abstract}
We present an initial investigation into Agentic Retrieval-Augmented Generation (RAG) for Ukrainian, conducted within the UNLP 2026 Shared Task on Multi-Domain Document Understanding. Our system combines two-stage retrieval (BGE-M3 with BGE reranking) with a lightweight agentic layer performing query rephrasing and answer-retry loops on top of Qwen2.5-3B-Instruct. Our analysis reveals that retrieval quality is the primary bottleneck: agentic retry mechanisms improve answer accuracy but the overall score remains constrained by document and page identification. We discuss practical limitations of offline agentic pipelines and outline directions for combining stronger retrieval with more advanced agentic reasoning for Ukrainian.

\end{abstract}

\section{Introduction}

Retrieval-Augmented Generation (RAG) has emerged as a key paradigm for grounding large language model (LLM) outputs based on external data \cite{lewis2020rag, gao2024ragsurvey}. Traditional RAG follows a fixed retrieve-then-generate pipeline, which often falls short when tasks require multi-step reasoning or iterative answer refinement. These limitations have motivated Agentic RAG, which embeds autonomous agent behaviors like planning, reflection, tool use, and multi-step retrieval \cite{singh2025agentic, li2025towards}.

The application of RAG to languages beyond English remains uneven. Despite its large speaker base and growing NLP ecosystem driven by the UNLP workshop series \cite{romanyshyn2024unlp}, agentic retrieval systems for Ukrainian have not been studied. The availability of multilingual models and embeddings has made basic RAG feasible for many languages, yet the agentic extensions that push the state of the art in English have not been explored for Ukrainian. The concept of agentic retrieval is closely related to established NLP problems: fact-checking systems require both retrieval and verification of factual correctness \cite{fact-checking}, while two-stage de-duplication systems mirror the retrieve-then-reason architecture central to RAG \cite{geodd}.

We present an initial investigation into Agentic RAG for Ukrainian within the UNLP 2026 Shared Task on Multi-Domain Document Understanding, which requires systems to answer multiple-choice questions and identify the supporting document and page\footnote{\url{https://github.com/unlp-workshop/unlp-2026-shared-task}}. The competition runs on Kaggle in code-only mode on a single P100 GPU with a 9-hour time limit, imposing realistic constraints on agentic pipeline design.

Our contributions are: (1)~a modular pipeline combining dense retrieval with reranking, open-weight LLMs, and agentic retry mechanisms; (2) ~discussion of practical challenges and future directions for agentic RAG for Ukrainian.

\section{Related Work}

RAG was introduced by \citet{lewis2020rag} to enhance LLM outputs with retrieved documents, and has been extended with adaptive retrieval and self-reflection mechanisms \cite{gao2024ragsurvey, asai2024selfrag}. Agentic RAG goes further by embedding autonomous agents into the pipeline. \citet{singh2025agentic} provide a taxonomy of architectures based on reflection, planning, tool use, and multi-agent collaboration. \citet{li2025towards} synthesize retrieval-reasoning interplay under a unified framework, identifying synergized approaches where agentic LLMs iteratively interleave search and reasoning.

Recent benchmarks have begun evaluating agentic RAG. \citet{xi2025infodeepseek} introduce InfoDeepSeek for assessing agentic information seeking in dynamic web environments, finding that search quality significantly impacts performance. \citet{ning2025searchmm} propose MC-Search for evaluating multimodal agentic RAG with structured reasoning chains. Both works highlight that retrieval quality remains a key bottleneck: a finding consistent with our results.

For Ukrainian NLP we can analyze, the UNLP 2024 Shared Task focused on LLM fine-tuning \cite{romanyshyn2024unlp}, with the winning system using RAG alongside fine-tuning \cite{boros2024sherlock}.

\section{Shared Task}

The UNLP 2026 Shared Task requires systems to: (1) predict the correct answer to a multiple-choice question given a document collection, and (2) identify the supporting document and page. The data spans multiple domains, with \texttt{dev}/\texttt{test\_public} containing two domains and \texttt{test\_private} containing three (including one unseen domain). The final score combines answer accuracy (\(a_i\)), document correctness (\(d_i\)), and page proximity (\(p_i\)):
\[
\mathrm{Metric}
=
0.5 \cdot \overline{a}
+
0.25 \cdot \overline{d}
+
0.25 \cdot \overline{p},
\]
where \(\overline{a}\), \(\overline{d}\), and \(\overline{p}\) denote the averages over all questions. Page proximity decreases linearly with page distance when the document is correct, and is zero otherwise.

\section{Methodology}

\subsection{Retrieval}

We evaluated sparse methods (TF-IDF with word, character, and hybrid n-grams) and dense retrieval (multilingual-e5-small, xlm-roberta-base, gte-multilingual-base, bge-m3). Character-level TF-IDF outperforms word-level, likely due to robustness to Ukrainian morphological variation. Our best configuration combines BGE-M3 for initial retrieval with BGE-reranker-v2-m3 for reranking.

\subsection{LLM-Based Answer Generation}

We evaluated five open-weight LLMs (Gemma, LapaLLM, Llama-2-7b, MamayLM, Qwen2.5-3B-Instruct) in LLM-only and LLM+retrieval settings. Qwen2.5-3B-Instruct performed best in both settings.

\subsection{Agentic Layer}

Our agentic layer implements two lightweight retry mechanisms: (1)~\textbf{Query rephrasing}: when model confidence is low, the agent rephrases the question and performs a second retrieval pass; (2)~\textbf{Answer retry}: the agent re-prompts the LLM with additional instructions, selecting the final answer based on confidence across attempts. Both are constrained by the single-GPU, 9-hour time limit.

\section{Evaluation}

\subsection{Retrieval Quality}

Table~\ref{tab:search_dev_metrics} reports retrieval performance on the development set. BGE-M3 with reranking achieves the best results (mean $d_i = 0.922$, mean $p_i = 0.811$), substantially outperforming all single-model approaches.

\begin{table}[t]
\centering
\caption{Search quality on the development set. Higher values indicate better retrieval.}
\label{tab:search_dev_metrics}
\resizebox{\columnwidth}{!}{
\begin{tabular}{lcc}
\toprule
\textbf{Method} & \textbf{Mean \(d_i\)} & \textbf{Mean \(p_i\)} \\
\midrule
tfidf\_word & 0.7202 & 0.5855 \\
tfidf\_char & 0.8742 & 0.7161 \\
tfidf\_hybrid & 0.8438 & 0.6878 \\
\midrule
multilingual-e5-small & 0.9109 & 0.7895 \\
xlm-roberta-base & 0.7202 & 0.5804 \\
gte-multilingual-base & 0.6074 & 0.4923 \\
bge-m3 & 0.9002 & 0.7365 \\
\midrule
bge-m3 + bge reranker & \textbf{0.9219} & \textbf{0.8111} \\
\bottomrule
\end{tabular}
}
\end{table}

\subsection{Answer Selection}

Table~\ref{tab:answer_metrics} compares LLM-only and LLM+retrieval performance. All models improve substantially with retrieved context, with Qwen2.5-3B-Instruct reaching 0.69 accuracy (+27 points over LLM-only). A simple answer-page similarity baseline achieves 0.49, outperforming all LLM-only configurations. Ukrainian-specific models (LapaLLM, MamayLM) perform comparably to multilingual baselines, suggesting that domain knowledge from retrieval is more important than language-specific pretraining.

\begin{table}[t]
\centering
\caption{Answer accuracy on the development set: LLM-only (no context) vs.\ LLM with retrieved context.}
\label{tab:answer_metrics}
\begin{tabular}{lcc}
\toprule
\textbf{Method} & \textbf{LLM-only} & \textbf{+Retrieval} \\
\midrule
Baseline (similarity) & --- & 0.49 \\
Gemma & 0.38 & 0.62 \\
LapaLLM & 0.35 & 0.66 \\
Llama-2-7b & 0.39 & 0.63 \\
MamayLM & 0.39 & 0.64 \\
Qwen2.5-3B-Instruct & \textbf{0.42} & \textbf{0.69} \\
\bottomrule
\end{tabular}
\end{table}

\subsection{Effect of Agentic Mechanisms}

Table~\ref{tab:agentic_metrics} shows the impact of agentic retry on the final metric (Qwen2.5-3B-Instruct + BGE-M3 with reranker). Combining both query rephrasing and answer retry achieves 0.7704, a gain of $\sim$1 point over the base pipeline. The improvements are consistent but modest, reflecting the constraints of our minimal agentic design compared to the sophisticated patterns described in recent surveys \cite{li2025towards, singh2025agentic}.

\begin{table}[t]
\centering
\caption{Effect of agentic mechanisms on the final metric (dev set, Qwen2.5-3B-Instruct + BGE-M3 reranker).}
\label{tab:agentic_metrics}
\begin{tabular}{lc}
\toprule
\textbf{Method} & \textbf{Final metric} \\
\midrule
LLM-only answering & 0.7607 \\
+ query rephrasing & 0.7615 \\
+ answer retry & 0.7629 \\
+ both & \textbf{0.7704} \\
\bottomrule
\end{tabular}
\end{table}

\subsection{Leaderboard Results and Analysis}

Table~\ref{tab:leaderboard} reports our two best submissions. The non-agentic pipeline achieves a private score of 0.7664 with strong retrieval ($d_i = 0.907$, $p_i = 0.814$) but moderate answer accuracy ($a_i = 0.633$). The agentic pipeline scores slightly higher (0.7668) with substantially better answer accuracy ($a_i = 0.814$) but lower page precision ($p_i = 0.625$), suggesting that query rephrasing retrieves relevant but different pages.

\begin{table}[t]
\centering
\caption{Leaderboard submissions: non-agentic vs.\ agentic.}
\label{tab:leaderboard}
\begin{tabular}{lcc}
\toprule
\textbf{Metric} & \textbf{Non-agentic} & \textbf{Agentic} \\
\midrule
Public score & 0.7365 & 0.7368 \\
Private score & 0.7664 & 0.7668 \\
\midrule
Mean \(a_i\) & 0.6334 & 0.8142 \\
Mean \(d_i\) & 0.9067 & 0.9110 \\
Mean \(p_i\) & 0.8142 & 0.6247 \\
Final metric & 0.7469 & 0.7437 \\
\bottomrule
\end{tabular}
\end{table}

Our results (12th place on the leaderboard) confirm several findings. First, \textbf{retrieval is the primary bottleneck}: the largest performance gains come from improving retrieval, not from changing the LLM or adding agentic mechanisms that consistent with InfoDeepSeek findings \cite{xi2025infodeepseek}. Second, \textbf{agentic mechanisms present a trade-off}: higher answer accuracy comes at the cost of page localization precision. Third, \textbf{small LLMs are effective with good retrieval}: the 27-point accuracy improvement from LLM-only to LLM+retrieval shows that retrieval quality matters more than model size for this task.

\section{Conclusion}

We have presented an initial investigation into Agentic RAG for Ukrainian, using the UNLP 2026 Shared Task as a testbed. Our results show that retrieval quality dominates performance, with agentic retry mechanisms providing modest but consistent improvements. The competition's offline, single-GPU constraints limit the agentic capabilities that can be explored; advanced patterns such as multi-agent collaboration and iterative deep search \cite{li2025towards, xi2025infodeepseek} remain out of reach in this setting.

Future work should address three directions: (1) improving retrieval through domain-adaptive embeddings and Ukrainian-specific rerankers; (2) exploring sophisticated agentic architectures (multi-step planning, self-reflection, chain-of-retrieval) in unconstrained settings; and (3) combining shared task insights with larger, API-accessible models to quantify the full potential of Agentic RAG for Ukrainian.

\section*{Limitations}

Our agentic capabilities are minimal (single-step retry) and do not capture the full spectrum of agentic behaviors \cite{singh2025agentic}. The single-GPU, 9-hour constraint precludes larger models, multi-agent architectures, or API-based services. Our evaluation is limited to the UNLP 2026 dataset, and we do not fine-tune models on Ukrainian data, relying on multilingual pretraining. Generalizability to other Ukrainian NLP tasks remains to be established.

\section*{Ethical Consideration}

We used Claude Opus 4.6 (Anthropic) and Grammarly for assistance with text editing and improving the writing of this paper.

\bibliography{custom}

@InProceedings{geodd,
author="Trokhymovych, Mykola
and Kosovan, Oleksandr",
editor="Silhavy, Radek
and Silhavy, Petr
and Prokopova, Zdenka",
title="GeoDD: End-to-End Spatial Data De-duplication System",
booktitle="Data Science and Algorithms in Systems",
year="2023",
publisher="Springer International Publishing",
address="Cham",
pages="717--727",
isbn="978-3-031-21438-7"
}

@inproceedings{fact-checking,
author = {Trokhymovych, Mykola and Saez-Trumper, Diego},
title = {WikiCheck: An End-to-end Open Source Automatic Fact-Checking API based on Wikipedia},
year = {2021},
isbn = {9781450384469},
publisher = {Association for Computing Machinery},
address = {New York, NY, USA},
url = {https://doi.org/10.1145/3459637.3481961},
doi = {10.1145/3459637.3481961},
booktitle = {Proceedings of the 30th ACM International Conference on Information \& Knowledge Management},
pages = {4155–4164},
numpages = {10},
location = {Virtual Event, Queensland, Australia},
series = {CIKM '21}
}

@inproceedings{lewis2020rag,
author = {Lewis, Patrick and Perez, Ethan and Piktus, Aleksandra and Petroni, Fabio and Karpukhin, Vladimir and Goyal, Naman and K\"{u}ttler, Heinrich and Lewis, Mike and Yih, Wen-tau and Rockt\"{a}schel, Tim and Riedel, Sebastian and Kiela, Douwe},
title = {Retrieval-augmented generation for knowledge-intensive NLP tasks},
year = {2020},
isbn = {9781713829546},
publisher = {Curran Associates Inc.},
address = {Red Hook, NY, USA},
booktitle = {Proceedings of the 34th International Conference on Neural Information Processing Systems},
articleno = {793},
numpages = {16},
location = {Vancouver, BC, Canada},
series = {NIPS '20}
}

@misc{gao2024ragsurvey,
      title={Retrieval-Augmented Generation for Large Language Models: A Survey}, 
      author={Yunfan Gao and Yun Xiong and Xinyu Gao and Kangxiang Jia and Jinliu Pan and Yuxi Bi and Yi Dai and Jiawei Sun and Meng Wang and Haofen Wang},
      year={2024},
      eprint={2312.10997},
      archivePrefix={arXiv},
      primaryClass={cs.CL},
      url={https://arxiv.org/abs/2312.10997}, 
}

@misc{singh2025agentic,
      title={Agentic Retrieval-Augmented Generation: A Survey on Agentic RAG}, 
      author={Aditi Singh and Abul Ehtesham and Saket Kumar and Tala Talaei Khoei},
      year={2025},
      eprint={2501.09136},
      archivePrefix={arXiv},
      primaryClass={cs.AI},
      url={https://arxiv.org/abs/2501.09136}, 
}

@misc{li2025towards,
      title={Towards Agentic RAG with Deep Reasoning: A Survey of RAG-Reasoning Systems in LLMs}, 
      author={Yangning Li and Weizhi Zhang and Yuyao Yang and Wei-Chieh Huang and Yaozu Wu and Junyu Luo and Yuanchen Bei and Henry Peng Zou and Xiao Luo and Yusheng Zhao and Chunkit Chan and Yankai Chen and Zhongfen Deng and Yinghui Li and Hai-Tao Zheng and Dongyuan Li and Renhe Jiang and Ming Zhang and Yangqiu Song and Philip S. Yu},
      year={2025},
      eprint={2507.09477},
      archivePrefix={arXiv},
      primaryClass={cs.CL},
      url={https://arxiv.org/abs/2507.09477}, 
}

@inproceedings{romanyshyn2024unlp,
    title = "The {UNLP} 2024 Shared Task on Fine-Tuning Large Language Models for {U}krainian",
    author = "Romanyshyn, Mariana  and
      Syvokon, Oleksiy  and
      Kyslyi, Roman",
    editor = "Romanyshyn, Mariana  and
      Romanyshyn, Nataliia  and
      Hlybovets, Andrii  and
      Ignatenko, Oleksii",
    booktitle = "Proceedings of the Third Ukrainian Natural Language Processing Workshop (UNLP) @ LREC-COLING 2024",
    month = may,
    year = "2024",
    address = "Torino, Italia",
    publisher = "ELRA and ICCL",
    url = "https://aclanthology.org/2024.unlp-1.9/",
    pages = "67--74",
}

@article{asai2024selfrag,
      author    = {Asai, Akari and Wu, Zeqiu and Wang, Yizhong and Sil, Avirup and Hajishirzi, Hannaneh},
      title     = {{Self-RAG}: Learning to Retrieve, Generate, and Critique through Self-Reflection},
      year      = {2023},
     journal    = {arXiv preprint arXiv:2310.11511},
     url        = {https://arxiv.org/abs/2310.11511}
    }

@misc{xi2025infodeepseek,
      title={InfoDeepSeek: Benchmarking Agentic Information Seeking for Retrieval-Augmented Generation}, 
      author={Yunjia Xi and Jianghao Lin and Menghui Zhu and Yongzhao Xiao and Zhuoying Ou and Jiaqi Liu and Tong Wan and Bo Chen and Weiwen Liu and Yasheng Wang and Ruiming Tang and Weinan Zhang and Yong Yu},
      year={2025},
      eprint={2505.15872},
      archivePrefix={arXiv},
      primaryClass={cs.IR},
      url={https://arxiv.org/abs/2505.15872}, 
}

@misc{ning2025searchmm,
      title={MC-Search: Evaluating and Enhancing Multimodal Agentic Search with Structured Long Reasoning Chains}, 
      author={Xuying Ning and Dongqi Fu and Tianxin Wei and Mengting Ai and Jiaru Zou and Ting-Wei Li and Hanghang Tong and Yada Zhu and Hendrik Hamann and Jingrui He},
      year={2026},
      eprint={2603.00873},
      archivePrefix={arXiv},
      primaryClass={cs.AI},
      url={https://arxiv.org/abs/2603.00873}, 
}

@inproceedings{boros2024sherlock,
    title = "Fine-Tuning and Retrieval Augmented Generation for Question Answering Using Affordable Large Language Models",
    author = "Boros, Tiberiu  and
      Chivereanu, Radu  and
      Dumitrescu, Stefan  and
      Purcaru, Octavian",
    editor = "Romanyshyn, Mariana  and
      Romanyshyn, Nataliia  and
      Hlybovets, Andrii  and
      Ignatenko, Oleksii",
    booktitle = "Proceedings of the Third Ukrainian Natural Language Processing Workshop (UNLP) @ LREC-COLING 2024",
    month = may,
    year = "2024",
    address = "Torino, Italia",
    publisher = "ELRA and ICCL",
    url = "https://aclanthology.org/2024.unlp-1.10/",
    pages = "75--82"
}

\end{document}